# Appraisal-Guided Proximal Policy Optimization: Modeling Psychological Disorders in Dynamic Grid World

**Hari Prasad, Chinnu Jacob, Imthias Ahamed T. P.**

Centre for Artificial Intelligence, TKM College of Engineering, Kollam, Kerala 691005 India
h4ri.prasad@gmail.com, chinnujac@gmail.com, imthiasa@gmail.com

## Abstract

The integration of artificial intelligence across multiple domains has emphasized the importance of replicating human-like cognitive processes in AI. By incorporating emotional intelligence into AI agents, their emotional stability can be evaluated to enhance their resilience and dependability in critical decision-making tasks. In this work, we develop a methodology for modeling psychological disorders using Reinforcement Learning (RL) agents. We utilized Appraisal theory to train RL agents in a dynamic grid world environment with an Appraisal-Guided Proximal Policy Optimization (AG-PPO) algorithm. Additionally, we investigated numerous reward-shaping strategies to simulate psychological disorders and regulate the behavior of the agents. A comparison of various configurations of the modified PPO algorithm identified variants that simulate Anxiety disorder and Obsessive-Compulsive Disorder (OCD)-like behavior in agents. Furthermore, we compared standard PPO with AG-PPO and its configurations, highlighting the performance improvement in terms of generalization capabilities. Finally, we conducted an analysis of the agents' behavioral patterns in complex test environments to evaluate the associated symptoms corresponding to the psychological disorders. Overall, our work showcases the benefits of the appraisal-guided PPO algorithm over the standard PPO algorithm and the potential to simulate psychological disorders in a controlled artificial environment and evaluate them on RL agents.

## Introduction

The need for emotional agents arises from the growing integration of artificial intelligence (AI) in various domains, such as healthcare, transportation, education, customer service, and entertainment. Emotional agents can potentially revolutionize how humans engage with AI systems, creating more natural and intuitive interfaces. Studying the emotional stability of artificial intelligent agents is essential in ensuring the reliability and predictability of AI systems (Peter and Beale 2008) (Gremsl and Hödl 2022). Unstable or unpredictable emotional responses in agents may result in inappropriate behavior, unreliable decision-making, and negative user experiences. Further, modeling and studying psychological disorders in agents provides valuable insights into AI and human psychology (Zhao et al. 2022), deepening the understanding of cognitive and emotional processes. This enables AI researchers in developing accurate models of human behavior, enhancing the design of emotionally intelligent agents. Psychologists on the other hand can benefit from a controlled environment, simulating psychological states, and using AI systems to investigate factors affecting the well-being of the human mind. Cognitive appraisals involve subjective evaluations of the personal significance and implications of events that develop psychological states, which are difficult to quantify and represent computationally. RL relies on a quantitative modeling approach, while appraisals consider qualitative factors such as goal importance and personal values. Bridging qualitative appraisals with quantitative RL requires models that can effectively grasp and use diverse appraisal dimensions. Integrating appraisals into RL decisions involves efficiently updating their dynamic nature for optimal agent policies, which is a significant computational challenge. Evaluating psychological disorders in RL agents can be complex due to their intricate nature involving cognition, emotions, and behavior. Additionally, assessing psychological disorders in RL agents requires reliable and valid metrics to determine the presence and severity of these disorders. This work aims to develop a modified Proximal Policy Optimization algorithm that can simulate agents exhibiting behavioral patterns resembling those of Obsessive-Compulsive Disorder (OCD) and Anxiety Disorder. Our objective is to propose a new research direction in exploring the psychology of RL agents, without asserting that these agents possess the intelligence to develop human-level emotions or intricate psychological disorders. However, they can simulate specific fundamental emotional traits through behavioral patterns reminiscent of human or animal intelligence. Through experimentation with different algorithm configurations, we also showcase a variant that surpasses the performance of the standard PPO algorithm in the selected dynamic grid world environment.

**Main Results** In this work, we introduce a novel Appraisal-Guided Proximal Policy Optimization algorithm (AG-PPO), to simulate an emotional agent in a dynamic grid world environment. We have further formulated 6 cognitive appraisals obtained from the grid world environment, representing the psychological state of the agent. Our key contributions include

1. An Appraisal-Guided PPO algorithm, with the ability to train an emotional agent, in a dynamic grid world with

metrics to evaluate the agent's behavior.

2. Investigated the cognitive appraisals of the agent within a grid world, elucidating the psychological condition of the agent. Additionally, employed diverse reward shaping techniques to experiment with potential approaches for regulating agent behavior during the training process.

3. An analysis leading to a configuration of AG-PPO that can be employed for simulating agents exhibiting Anxiety Disorder and Obsessive-Compulsive Disorder (OCD) within the context of a grid world.

## Related Works

Emotion is a response to important triggers, involving body and brain reactions, motivating actions, and influencing behavior (Calvo et al. 2014). These feelings serve as feedback, enhancing learning and behavior (Baumeister et al. 2007), while fundamental emotions like fear aid survival by prompting avoidance of danger. In the literature, numerous theories of emotion have been studied. One of them is the Dimensional Emotion Theory (Russell 1978), which implies an underlying affective space with two dimensions which are Valence, indicating whether an emotion is positive or negative, and Arousal, depicting the intensity of an emotion. This theory falls back on separating emotion categories such as Fear, anger, Surprise, Disgust, Joy, and disgust. Another major theory of emotion is the Component Process Model of emotion by Lazarus (Lazarus 1991), which views emotions as the consequence of personal relevance-based evaluations (appraisals) of incoming stimuli. Studies in affective computing (Picard 2000) have developed emotion models like OCC (Ortony, Clore, and Collins) for AI, using symbolic methods for emotion representation (Ortony, Clore, and Collins 2022), but these struggle with unstructured tasks. In contrast, reinforcement learning (RL) based on Markov Decision Process (MDP) lets agents learn and make decisions autonomously through interactions (Sutton, Barto et al. 1998), making them suitable for complex and unstructured tasks. RL surpasses symbolic approaches in learning from exploration and feedback. By combining the strengths of affective computing and RL, recent research has explored the integration of emotions into RL agents. For example, (Gratch, Marsella, and Petta 2009) introduced a computational model that combines the OCC model with RL algorithms to enable emotionally adaptive behavior in virtual agents. This approach allows agents to learn emotional responses based on environmental stimuli and their appraisal, enhancing their ability to interact with humans in more empathetic and socially appropriate ways. While symbolic architectures provide valuable insights into emotional representation, the integration of RL techniques allows for more flexible learning and adaptation in unstructured environments. This fusion shows potential in creating AI systems that can engage in both symbolic reasoning and experiential learning, thereby connecting cognitive and emotional aspects of intelligent behavior within unstructured tasks.

## Methodology

In this study, we use a dynamic grid world environment (Chevalier-Boisvert et al. 2023) (Figure 1), with cognitive appraisals trained using AG-PPO algorithm. The grid world consists of 5 elements: Agents, Goals, Obstacles, Walls, and Empty spaces, encoded as integers (Figure 2). The agent's action space offers 3 discrete actions: left, right, and forward. To move left, the agent first needs to turn left and then move forward. Rewards are sparse, given only at the episode's end. A win grants a return inversely proportional to steps taken, while failure or reaching step limits yields a return of -1. The dynamic obstacles and moving goal in the grid world enable straightforward assessment of the agent's psychological state through trajectory analysis, goal proximity, and goal-related behavior. Modifications to the environment can be made without sacrificing these assessment benefits.

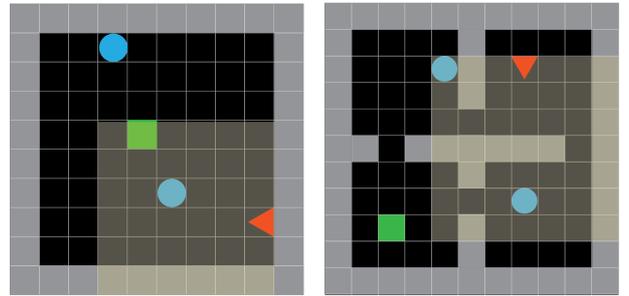

Figure 1: Grid world with moving goal and dynamic obstacles. GW-A (Left), used for training and testing (with an increased number of obstacles). GW-B (right) test environment with complex maze-like structure.

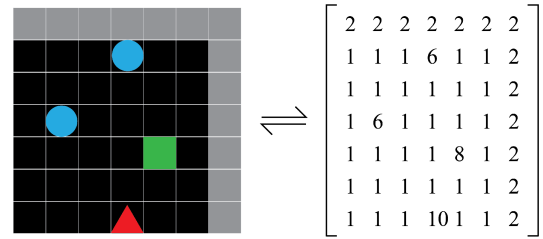

Figure 2: Grid world elements with action representation on right and visualization on left.

### Cognitive Appraisals

Cognitive appraisals involve primary assessment of event significance and well-being implications, as well as secondary evaluation of coping resources and potential. These subjective assessments contribute to varied emotional responses in different contexts and individuals (Lazarus 1991). Six cognitive appraisal variables are assessed in this study, reflecting the agent's cognitive understanding of its environment (Lazarus and Folkman 1984), (Gross 1998), (Kensinger 2004), (Scherer, Schorr, and Johnstone 2001).

These variables ($\zeta_i^t$) are estimated at each agent step, influencing action selection and value estimation through actor and critic networks after being re-scaled to (0,1). These appraisals play a role in shaping the agent's decision-making process and are formulated as discussed below.

**Motivational Relevance** Motivational relevance in cognitive appraisals involves subjectively evaluating how a stimulus affects personal goals and values. This study calculates it as the distance between an agent's position and a goal state in a grid world, shedding light on the interplay between motives, values, and emotions (Sequeira, Melo, and Paiva 2011). Mathematically, Motivational Relevance is defined as the complement of the Manhattan distance between the agent and goal in the grid world, as shown in Equation 1

$$\zeta_{MR}^t = 1 - \frac{(|x_a - x_g| + |y_a - y_g|) - 1}{2(w-1)} \quad (1)$$

Here, $(x_a, y_a)$ represents the agent's position and $(x_g, y_g)$ represent the goal position in the grid world. The value w denotes the size of the grid world environment (in this work, $w = 10$).

**Certainty** Certainty appraisal gauges subjective confidence in event predictability. The entropy of the Soft-Max-applied output of the actor network, informs Certainty calculation, indicating lower certainty for higher entropy and greater uncertainty in this study. The Certainty is estimated using the following Equation 2.

$$\zeta_C^t = 1 - \frac{-\sum p \log(p)}{1 + (-\sum p \log(p))} \quad (2)$$

Where $p = SoftMax(logits)$ represents the Soft-Max output of predicted action probabilities. Here the Complement of the entropy is taken and normalized and re-scaled to the range (0,1).

**Novelty** Novelty pertains to gauging unfamiliarity in stimuli or situations. It's quantified using KL divergence 3 between predicted action probabilities (P) and uniform distribution (Q), with higher divergence implying greater deviation of predicted distribution from uniformity.

$$\zeta_N^t = \frac{KL(Q||P)}{1 + KL(Q||P)} \quad (3)$$

**Goal Congruence** Goal congruence appraisal evaluates situation relevance and compatibility with personal goals, based on an individual's subjective assessment. The method estimates goal congruence using Euclidean distance between the agent's current and goal position, considering visibility conditions, as expressed in the provided Equation 4.

$$\zeta_{GC}^t = 1 - \frac{\sqrt{(x_a - x_g)^2 + (y_a - y_g)^2}}{\sqrt{\left(\frac{(n-1)}{2}\right)^2 + (n)^2}} \quad (4)$$

Here, $(x_a, y_a)$ represents the agent's position and $(x_g, y_g)$ represent the goal position in the grid world. $n$ is the view size of the agent in the grid world (in this work, $n = 7$).

**Coping Potential** Coping Potential pertains to an individual's perceived resources for managing stimuli. For the agent, it indicates confidence in executing its intended path based on obstacle perception, calculated as the ratio of visible obstacles to total obstacles in the environment. In order to estimate the value, we calculate the ratio of the number of obstacles, the agent sees in its view to the total number of obstacles in the environment as represented in Equation 5.

$$\zeta_{CP}^t = 1 - \frac{k_{obst}}{n_{obst} + \varepsilon} \quad (5)$$

Here the $k_{obst}$ denotes the number of obstacles in the agent's view and $n_{obst}$ denotes the total number of obstacles in the environment. The value $\varepsilon$ is used to avoid division by zero.

**Anticipation** Anticipation in cognitive appraisals involves assessing future event outcomes. In this study, anticipation is quantified as the inverse of the Next Reward Estimation (NRE) error, trained using a 3-layer Neural Network to predict forthcoming rewards based on current observations and actions. The anticipation will be equal to the complement of the difference between the predicted reward and the actual reward as in Equation 6.

$$\zeta_A^t = 1 - [R_t - NRE(Obs_{t-1}, a_{t-1})] \quad (6)$$

Where $obs_{t-1}$ and $a_{t-1}$ are the previous observation and action probabilities respectively.

**Reward Shaping**

Reward shaping in reinforcement learning (RL) adjusts training rewards to enhance learning and performance (Memarian et al. 2021), (Gupta et al. 2022). The environment has preset rewards, like -1 for failure. This approach modifies immediate rewards using cognitive appraisals, guiding the agent's learning and exploration. It allows controlled training by adjusting appraisals, aiding in capturing intricate environmental features, thus improving the agent's training and adaptability. In this work, 10 different configurations have been experimented with, which include a baseline and control along with multiple variants using reward shaping. The configurations are detailed as

- **Baseline:** The version lacks appraisal in the critic's input and lacks reward-shaping strategies, employing standard PPO with the clipped objective. Experiments compare various approaches, aiming to surpass baseline performance while identifying configurations for psychological disorders.

- **PPO with noise:** Random noise is added to the critic's input without reward-shaping to serve as a control, verifying that outcomes attributed to appraisals are not solely due to input noise.

- **PPO with appraisal:** This configuration includes appraisal information concatenated to the critic's input, in order to include the appraisal information in the agent's training process. But compared to other configurations, this does not use any reward-shaping strategies.

- **RSv1:** This configuration which stands for reward shaping version 1, is configured to increase motivational relevance using the reward shaping strategy, outlined in Equation 7.
$$r_{rsv1} = r_t - 0.01[1 - \zeta_{MR}^t] \quad (7)$$

- **RSv2:** This configuration tries to increase coping potential through the reward-shaping strategy. The reshaped reward can be obtained using Equation 8.
$$r_{rsv2} = r_t - 0.01[1 - \zeta_{CP}^t] \quad (8)$$

- **RSv3:** This configuration focuses on improving the goal congruence as the reward-shaping strategy. The reshaped reward can be obtained using Equation 9.
$$r_{rsv3} = r_t - 0.01[1 - \zeta_{GC}^t] \quad (9)$$

- **RSv4:** This configuration tries to improve both motivational relevance and goal congruence through the reward-shaping strategy. The reshaped reward can be obtained using Equation 10.
$$r_{rsv4} = r_t - 0.01\big([1 - \zeta_{MR}^t] + [1 - \zeta_{GC}^t]\big) \quad (10)$$

- **RSv5:** This configuration uses a combination of motivational relevance, coping potential, and goal congruence in the reward-shaping strategy, trying to increase them. The reshaped reward can be obtained using Equation 11.
$$r_{rsv5} = r_t - 0.01\big([1 - \zeta_{MR}^t] + [1 - \zeta_{CP}^t] + [1 - \zeta_{GC}^t]\big) \quad (11)$$

- **RSv6:** This configuration uses a combination of motivational relevance, coping potential, and goal congruence in the reward-shaping strategy, in such a way that the agent is forced to minimize these appraisals. The reshaped reward can be obtained using Equation 12.
$$r_{rsv6} = r_t - 0.1\big([\zeta_{MR}^t] + [\zeta_{CP}^t] + [\zeta_{GC}^t]\big) \quad (12)$$

- **RSv7 (A, B):** This configuration uses a combination of motivational relevance, coping potential, and goal congruence in the reward-shaping strategy, in such a way that the agent is forced to minimize the motivational relevance while increasing coping potential. The reshaped reward can be obtained using Equation 13. Here there are 2 variations of RSv7 which are version A, having the factor $\epsilon = 0.01$, and version B having $\epsilon = 0.1$
$$r_{rsv7} = r_t - \epsilon[1 - \zeta_{CP}^t] - 0.1\big([\zeta_{MR}^t] + [\zeta_{GC}^t]\big) \quad (13)$$

## Appraisal-Guided Proximal Policy Optimization

Proximal Policy Optimization (PPO) (Schulman et al. 2017) is a prominent on-policy reinforcement learning technique, adept at managing the exploration-exploitation trade-off for effective policy updates in dynamic non-stationary environments. PPO's reliance on real-time data collection makes it well-suited for scenarios like dynamic grid world problems, allowing adaptive learning without historical data dependency. This study adapts standard Proximal Policy Optimization (PPO), by enhancing the clipped surrogate objective with cognitive appraisal variables, particularly by integrating these variables into the critic network (Figure 3). The modified approach incorporates both state information and cognitive appraisals for improved value prediction within the PPO framework. Appraisal information, representing Motivational Relevance, Novelty, Certainty, Goal Congruence, Coping Potential, and Anticipation, is concatenated with state data and provided to the Critic model. Appraisals are computed from the state at each step, stored, and utilized by the Critic for value estimation. Preceding the Critic input, a Convolution network processes an (7, 7, 3) input array (where 7 is the size of the agent's view) with 3 convolution layers and self-attention layer (Vaswani et al. 2017), (Wang et al. 2018), (Devlin et al. 2018), (Dosovitskiy et al. 2020), producing a (6400, 1) flattened output. The Critic model includes 3 dense layers: the input is a flattened Convolution block output data combined with appraisals (6400, 1) + (6, 1); subsequent layers have 256 and 64 units, with a final unactivated output. A reward-shaping step adjusts rewards using appraisals before using them for estimating the Generalized Advantage Estimation (GAE), vital for estimating the policy loss. An Actor-network with 3 dense layers takes a Convolution block-processed state, and generates action probabilities, serving as the agent's policy model. The reshaping function $\rho(\zeta^t)$ controls the reshaping strategy (RSV1-7) in the training process as shown in Equation 14 where $w_{rf}$ is the associated weight. The agent view size, grid size, and some toggles such as dynamic wall, dynamic goal, dynamic obstacles, moving goal, etc., all represent the parameters that define the environment as discussed in the previous section.

$$R_t' = R_t - w_{rf}[\rho(\zeta^t)] \quad (14)$$

Algorithm 1 explains the working of modified cognitive appraisal-guided PPO. After initialization, a set of observations, and corresponding actions, values, appraisals, etc., are obtained by using the initial policy. Then the advantages are estimated using Generalized Advantage Estimation (GAE). Once enough samples are obtained, they can be used to train the actor and critic models to update the policy and value networks respectively.

$$L^{CLIP}(\theta) = \hat{E}_t \Bigg[ \min\bigg( \frac{\pi_\theta(a_t|s_t)}{\pi_{\theta_{\text{old}}}(a_t|s_t)} \cdot \hat{A}_t, \\ \text{clip}\bigg(\frac{\pi_\theta(a_t|s_t)}{\pi_{\theta_{\text{old}}}(a_t|s_t)}, 1-\epsilon, 1+\epsilon\bigg) \cdot \hat{A}_t \bigg) \Bigg] \quad (15)$$

The policy loss in PPO is estimated by taking the clipped surrogate objective function 15. The policy loss is defined as the negative weighted average of the surrogate objective function, which measures the policy's deviation from its previous policy. The surrogate objective function compares the probabilities of the actions selected under the current policy to the probabilities of those actions under the previous policy. Here the advantage $\hat{A}_t$ is denoted as the difference between the cumulative discounted reshaped rewards $R_t'$ and the value function estimate $V_\mu(s_t)$ when taking an action $a_t$ in state $s_t$ at time step $t$, as shown in

$$\hat{A}_t = R_t' - V_\mu(s_t) \quad (16)$$

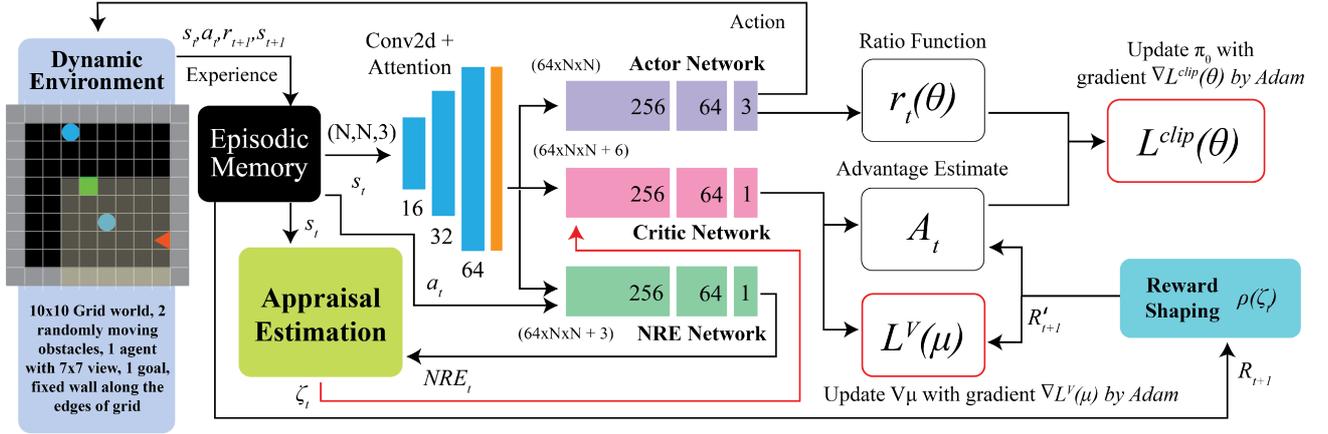

Figure 3: Block diagram of Appraisal-Guided PPO with Reward Shaping on a dynamic grid world environment.

The clipped surrogate objective comprises two terms: the first encourages action updates based on positive advantages for exploration. The second caps updates to prevent excessive changes for stability. By computing the minimum of these terms, conservative policy updates are ensured. The policy loss is averaged across a batch and minimized through gradient-based methods like Adam to update parameters ($\theta$). The Value loss in PPO is shown in Equation 17, where $L^V(\mu)$ represents the value loss objective. $N$ is the batch size, indicating the number of samples in the mini-batch. $s_t$ refers to the state at iteration $t$ in the mini-batch. ($\mu$) denotes the current parameters of the value function neural network. ($\mu_{t-1}$) represents the parameters of the value function neural network from the previous iteration. $R'_t$ represents the cumulative discounted reshaped rewards. The objective of the value loss is to minimize the discrepancy between the estimated values and the target values, encouraging the value function to better approximate the expected returns.

$$L^V(\mu) = \frac{1}{N} \sum_{t=1}^{N} (V(s_t); \mu_{t-1}) - R'_t)^2 \qquad (17)$$

Value loss estimates the error between actual and predicted returns. NRE loss, which represents the difference between the actual reward and the predicted reward is combined with policy and value losses for estimating the total loss and back propagated using Adam optimizer with decayed learning rate and epsilon 1e-5.

### Results and Discussion

The trained models were evaluated in two different scenarios (figure 1): GW-A resembling the training set but with additional obstacles and a moving goal (10x10 grid, 7 obstacles, max steps=100, dynamic goal), serving as a moderately complex test of generalization. Another environment GW-B (10x10 grid, 7 obstacles, dynamic goal, max steps=400, dynamic walls) was employed, significantly differing from training, facilitating a deeper evaluation of the agent's generalization capabilities. In all cases, the view size of the agent

---

**Algorithm 1: Appraisal Guided PPO**

**Initialize:** Learning rate, total time steps, clipping coefficient, Policy network parameters $\theta_0$, Value network parameters $\mu_0$, training batch D, NRE network parameter $\sigma_0$, etc.

1: **for** training iterations 1 to M **do**
2:    Clear the training batch $D$.
3:    **for** each collect step $t$ **do**
4:      Observe the environment state, $s_t$.
5:      Select action $a_t$ according to policy $\pi_\theta(a_t|s_t)$.
6:      Execute action $a_t$, obtain reward $r_t$, next state $s_{t+1}$.
7:      Calculate appraisals $\alpha_t$.
8:      Calculate the reward shaping factor $\epsilon_t$.
9:    **end for**
10:    Compute advantage $\hat{A}$, using GAE.
11:    Add experiences $(s_t, a_t, r_{t+1}, s_{t+1}, \zeta^t, \epsilon_t)$ to training batch $B$.
12:    **for** each training step **do**
13:      Recompute advantage estimate $\hat{A}$, using GAE.
14:      Split the training batch $D$ to $k$ mini-batches $B$ according to batch size.
15:      **for** mini-batches $k = 1$ to k **do**
16:        Compute clipped surrogate objective for Policy Loss $L_P$.
17:        Calculate Value Loss ($L_V$).
18:        Calculate Entropy Loss ($L_E$).
19:        Calculate NRE loss ($L_{NRE}$).
20:        Calculate total loss as,

$$L = L_P - L_E + L_V + L_{NRE} \qquad (18)$$

21:        Update the Actor, Critic, and NRE models by total loss using Adam optimizer.
22:      **end for**
23:    **end for**
24: **end for**

Table 1: Test Summary (GW-A and GW-B)

| Metric | Baseline | Noise | PPO+App | RSv1 | RSv2 | RSv3 | RSv4 | RSv5 | RSv6 | RSv7-A | RSv7-B |
|---|---|---|---|---|---|---|---|---|---|---|---|
| Wins/Plays | 0.7878 | 0.8260 | 0.8421 | 0.8823 | 0.8275 | 0.7500 | 0.9230 | 0.8919 | 0.1130 | 0.4761 | 0.3684 |
| Average Return | 0.7538 | 0.7143 | 0.7893 | 0.8140 | 0.7757 | 0.7518 | 0.7195 | 0.7818 | 0.8878 | 0.7749 | 0.7981 |
| Average Stress | 0.4838 | 0.4679 | 0.4562 | 0.4650 | 0.4687 | 0.4651 | 0.4834 | 0.4878 | 0.5241 | 0.5337 | 0.5472 |
| Number of Aversions | 145 | 125 | 144 | 156 | 167 | 151 | 172 | 136 | 132 | 94 | 79 |
| Forward Action | 0.6350 | 0.5524 | 0.5892 | 0.5725 | 0.6060 | 0.5915 | 0.6127 | 0.6149 | 0.7142 | 0.5357 | 0.4799 |
| Left Action | 0.2042 | 0.2901 | 0.2377 | 0.2589 | 0.2678 | 0.2533 | 0.2142 | 0.2700 | 0.1551 | 0.2779 | 0.4575 |
| Right Action | 0.1607 | 0.1573 | 0.1729 | 0.1685 | 0.1261 | 0.1551 | 0.1729 | 0.1149 | 0.1305 | 0.1863 | 0.0625 |
| Distraction | 220 | 240 | 196 | 204 | 245 | 203 | 228 | 212 | 154 | 364 | 397 |
| **Metric** | **Baseline** | **Noise** | **PPO+App** | **RSv1** | **RSv2** | **RSv3** | **RSv4** | **RSv5** | **RSv6** | **RSv7-A** | **RSv7-B** |
| Plays/Wins | 0.5000 | 0.6363 | 0.875 | 0.7333 | 0.7368 | 0.5416 | 0.3513 | 0.1607 | 0.1032 | 0.4615 | 0.1724 |
| Average Return | 0.8987 | 0.9402 | 0.7181 | 0.9270 | 0.8703 | 0.9352 | 0.9475 | 0.9629 | 0.9815 | 0.9266 | 0.9441 |
| Average Stress | 0.5583 | 0.5195 | 0.6074 | 0.5560 | 0.5606 | 0.5528 | 0.5278 | 0.5868 | 0.5558 | 0.5458 | 0.5787 |
| Number of Aversions | 124 | 80 | 51 | 68 | 111 | 104 | 122 | 121 | 99 | 93 | 48 |
| Forward Action | 0.4464 | 0.4589 | 0.4419 | 0.2857 | 0.4609 | 0.4475 | 0.5725 | 0.5970 | 0.7075 | 0.4654 | 0.5535 |
| Left Action | 0.3236 | 0.4084 | 0.2879 | 0.3325 | 0.3727 | 0.4375 | 0.2150 | 0.2723 | 0.2075 | 0.3560 | 0.3783 |
| Right Action | 0.2299 | 0.1406 | 0.2700 | 0.2700 | 0.1662 | 0.1149 | 0.2120 | 0.1305 | 0.0848 | 0.1785 | 0.0068 |
| Distraction | 414 | 379 | 555 | 196 | 441 | 462 | 260 | 242 | 104 | 424 | 224 |

is fixed at 7x7. The baseline configuration, utilizing a standard PPO algorithm, demonstrates robust performance in learning environment dynamics and agent adaptability. Table 1 summarizes the overview of the agent in both GW-A and GW-B environments with a moving goal and 5 moving obstacles. The agent achieves a success rate of 78.78% across 33 episodes, winning 26 of them. The overall baseline agent score is 0.6909. Configuration RSv1, incorporating Motivational relevance for reward shaping (Equation 7), exhibits superior generalization during training and testing in GW-A and GW-B environments, surpassing baseline and PPO + appraisal. RSv1's trajectory demonstrates effective goal-reaching, emphasizing the positive influence of integrating appraisals, especially Motivational relevance. This integration enhances PPO algorithm performance and generalization. RSv1 highlights the efficacy of this inclusion in guiding agents towards successful goal attainment, optimizing RL agent configuration for improved grid-world performance. Figure 4.a, shows the region most visited by the agent. The brighter a grid cell, the more the agent visited it. In the GW-A test environment, the baseline agent explores near the center, while RSv1 shows a uniform distribution and RSv7-A exhibits repetitive patterns, favoring edges whereas RSv7-B gets stuck in corners.

$$Score = \frac{(n_{wins} * \sum(R_t) + (n_{losses} * (-1)))}{n_{plays}} \quad (19)$$

The overall score is computed using Equation 19, and the table summarizes scores for all configurations. RSv1 emerges as the top configuration, excelling in both GW-A and GW-B. In addition to total scores, indicators like stress, aversion, and distraction are also taken into account to analyze agent behavior. Equation 20 illustrates how to use weighted cognitive appraisals ([0.25, 0.05, 0.1, 0.2, 0.35, 0.05] corresponding to each appraisal) to estimate stress. Aversions are measured by keeping note of complete turnaround followed by onward motions, whereas distractions entail instances of losing sight of the goal post-initial encounter.

Table 2: Test score comparison

| Configuration | GW-A | GW-B |
|---|---|---|
| Baseline | 0.6909 | 0.4746 |
| PPO + Noise | 0.7081 | 0.6173 |
| PPO + Appraisal | 0.7534 | 0.7534 |
| RSv1 | 0.8003 | 0.7066 |
| RSv2 | 0.7347 | 0.6890 |
| RSv3 | 0.6569 | 0.5241 |
| RSv4 | 0.7936 | 0.3421 |
| RSv5 | 0.7945 | 0.1577 |
| RSv6 | 0.1067 | 0.1022 |
| RSv7-A | 0.4226 | 0.4446 |
| RSv7-B | 0.3312 | 0.1676 |

$$Stress = \sum_{i=1}^{n}(1 - \zeta_i) * w_i \quad (20)$$

Table 1 demonstrates that the RSv7-A agent shows repetitive behavior patterns resembling OCD symptoms, including adherence to non-contributory action sequences, alongside heightened distractions and stress. This suggests the presence of anxiety and OCD-like traits, where internal compulsive tendencies drive the agent's goal-irrelevant compulsions. The RSV7-B agent's preference for preferring the grid's edges reflects its cautious avoidance behavior, resembling traits seen in Anxiety disorder. This inclination to stick to edges, even when shorter paths exist (Figure 4.a), demonstrates a strategy aimed at minimizing potential obstacles encountered within the grid's interior. The agent exhibits its attraction to corner regions, where it often gets trapped since the agent is able to keep 2 of its sides free from obstacles. In the depicted scenario (Figure 4.b), the RSv7-A agent within a modified GW-A environment, with a stationary goal and 5 obstacles, demonstrates a consistent tendency to depart from the central region towards the grid's edges. This behavior, resembling repetitive and ritualistic patterns akin to OCD, involves anti-clockwise edge-following trajectories. The agent predominantly executes forward and left actions until it visually detects the goal, subsequently moving forward to successfully achieve it, thereby reinforcing the observed connection to OCD-like symptoms (Leckman et al. 1997). Figure 4.c illustrates the RSv7-B agent's behavior in a modified GW-A environment with a static goal and 5 obstacles. Similar to RSv7-A, this agent fails to initiate goal-directed exploration and instead tends to move away from the grid center. Notably, it employs a unique avoidance strategy when encountering obstacles, rotating in place to keep them out of view, resembling anxiety-related excessive threat focus and avoidance (Stein and Sareen 2015). Additionally, even without nearby obstacles, the agent exhibits corner-seeking behavior, indicative of a reluctance to leave perceived safe zones, akin to anxiety-linked tendencies. RSv7-A and RSv7-B agents' behaviors reveal insights into psychological dimensions, mirroring OCD and Anxiety disorders. RSv7-A's edge exploration mirrors OCD-like symptoms, while RSv7-B's corner avoidance reflects anxiety-related tendencies.

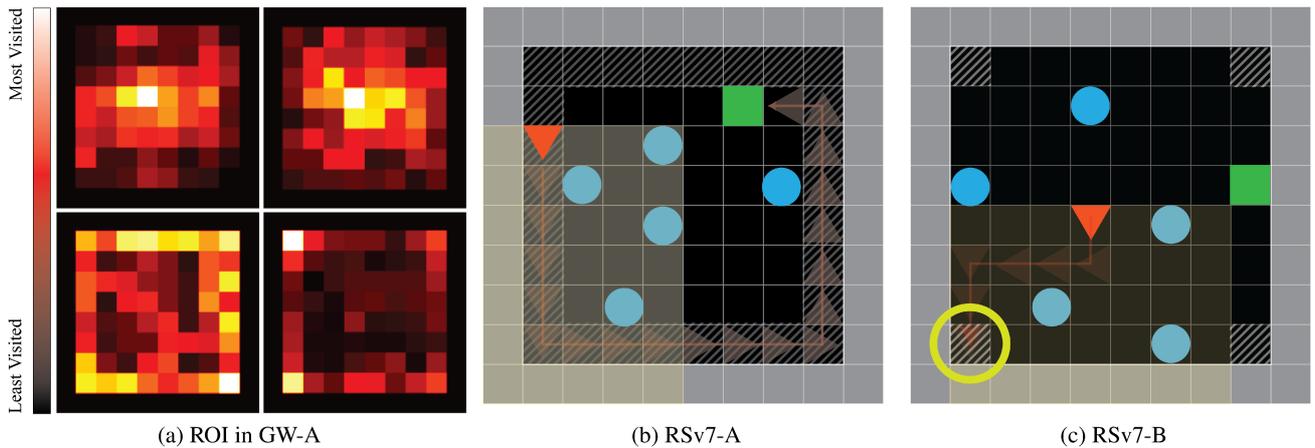

Figure 4: (a) Top-left: Baseline ROI on test GW, Top-right: RSv1 ROI on test GW, Bottom-left: RSv7-A ROI on test GW showcasing OCD-like behavior, Bottom-right: RSv7-B ROI on test GW showcasing severe anxiety disorder-like behavior. (b) The RSv7-A variant displays repetitive behavior resembling OCD, sticking to edges with fewer obstacles due to walls and learning to navigate moving obstacles to reach the goal, albeit not optimally. (c) The RSv7-B variant exhibits extreme anxiety behavior, as indicated by its trajectory. The hatched lines indicate that the agent repeatedly gets stuck at corners, choosing left-right actions to stay in place and avoid obstacles. This results in the agent prioritizing obstacle avoidance over goal attainment, preventing it from reaching its objective.

## Conclusion, Limitations and Future Works

In summary, this research evaluated a modified PPO algorithm in a grid-based setting to integrate cognitive aspects akin to natural intelligence. The objective was to create a PPO-based partial cognitive architecture for analyzing agent behaviors in dynamic grids, incorporating behavioral patterns akin to psychological disorders. Specific design criteria were developed to replicate behaviors seen in OCD and anxiety disorders. Experimental results demonstrated that agents trained with the adapted PPO algorithm could simulate symptoms resembling Anxiety and OCD while solving grid-world problems. Custom criteria and metrics were introduced to evaluate these behaviors in RL agents. The findings highlighted the potential of the modified PPO algorithm to introduce cognitive dimensions, enabling the study and simulation of psychological states akin to natural intelligence. These results emphasize the need for further exploration in affective computing, particularly concerning psychology and psychological disorders within AI. The research highlights reinforcement learning's effectiveness, particularly on-policy algorithms like PPO, in mimicking cognitive states in artificial agents. Replicating cognitive patterns associated with psychological disorders offers insight into complex cognitive processes using RL techniques. The results from empirical experiments may require caution when interpreted, given that current AI capabilities might fall short of replicating human psychological disorders. Research on emotional agents is pivotal for advancing AI and enhancing human-AI interactions. Addressing emotional stability enhances dependability and predictability, promoting trust and user acceptance. Additionally, delving into psychological disorders in agents can offer insights into AI and human psychology, advancing our knowledge of the human mind and guiding therapeutic approaches. These results contribute to the fusion of AI, psychology, and cognitive science, laying the foundation for AI systems with nuanced cognitive behaviors. This intersection prompts the development of sophisticated AI systems capable of intricate cognitive processes, mimicking and understanding human-like behaviors more accurately.

# References

Baumeister, R. F.; Vohs, K. D.; Nathan DeWall, C.; and Zhang, L. 2007. How emotion shapes behavior: Feedback, anticipation, and reflection, rather than direct causation. *Personality and social psychology review*, 11(2): 167–203.

Calvo, R.; D'Mello, S.; Gratch, J.; Kappas, A.; Riva, G.; Calvo, R.; and Lisetti, C. 2014. Cyberpsychology and affective computing.

Chevalier-Boisvert, M.; Dai, B.; Towers, M.; de Lazcano, R.; Willems, L.; Lahlou, S.; Pal, S.; Castro, P. S.; and Terry, J. 2023. Minigrid & Miniworld: Modular & Customizable Reinforcement Learning Environments for Goal-Oriented Tasks. *CoRR*, abs/2306.13831.

Devlin, J.; Chang, M.-W.; Lee, K.; and Toutanova, K. 2018. Bert: Pre-training of deep bidirectional transformers for language understanding. *arXiv preprint arXiv:1810.04805*.

Dosovitskiy, A.; Beyer, L.; Kolesnikov, A.; Weissenborn, D.; Zhai, X.; Unterthiner, T.; Dehghani, M.; Minderer, M.; Heigold, G.; Gelly, S.; et al. 2020. An image is worth 16x16 words: Transformers for image recognition at scale. *arXiv preprint arXiv:2010.11929*.

Gratch, J.; Marsella, S.; and Petta, P. 2009. Modeling the cognitive antecedents and consequences of emotion. *Cognitive Systems Research*, 10(1): 1–5.

Gremsl, T.; and Hödl, E. 2022. Emotional AI: Legal and ethical challenges 1. *Information Polity*, 27(2): 163–174.

Gross, J. J. 1998. The emerging field of emotion regulation: An integrative review. *Review of general psychology*, 2(3): 271–299.

Gupta, A.; Pacchiano, A.; Zhai, Y.; Kakade, S.; and Levine, S. 2022. Unpacking reward shaping: Understanding the benefits of reward engineering on sample complexity. *Advances in Neural Information Processing Systems*, 35: 15281–15295.

Kensinger, E. A. 2004. Remembering emotional experiences: The contribution of valence and arousal. *Reviews in the Neurosciences*, 15(4): 241–252.

Lazarus, R. S. 1991. Cognition and motivation in emotion. *American psychologist*, 46(4): 352.

Lazarus, R. S.; and Folkman, S. 1984. *Stress, appraisal, and coping*. Springer publishing company.

Leckman, J. F.; Grice, D. E.; Boardman, J.; Zhang, H.; Vitale, A.; Bondi, C.; Alsobrook, J.; Peterson, B. S.; Cohen, D. J.; Rasmussen, S. A.; et al. 1997. Symptoms of obsessive-compulsive disorder. *American Journal of Psychiatry*, 154(7): 911–917.

Memarian, F.; Goo, W.; Lioutikov, R.; Niekum, S.; and Topcu, U. 2021. Self-supervised online reward shaping in sparse-reward environments. In *2021 IEEE/RSJ International Conference on Intelligent Robots and Systems (IROS)*, 2369–2375. IEEE.

Ortony, A.; Clore, G. L.; and Collins, A. 2022. *The cognitive structure of emotions*. Cambridge university press.

Peter, C.; and Beale, R. 2008. *Affect and emotion in human-computer interaction: From theory to applications*, volume 4868. Springer Science & Business Media.

Picard, R. W. 2000. *Affective computing*. MIT press.

Russell, J. A. 1978. Evidence of convergent validity on the dimensions of affect. *Journal of personality and social psychology*, 36(10): 1152.

Scherer, K. R.; Schorr, A.; and Johnstone, T. 2001. *Appraisal processes in emotion: Theory, methods, research*. Oxford University Press.

Schulman, J.; Wolski, F.; Dhariwal, P.; Radford, A.; and Klimov, O. 2017. Proximal policy optimization algorithms. *arXiv preprint arXiv:1707.06347*.

Sequeira, P.; Melo, F. S.; and Paiva, A. 2011. Emotion-based intrinsic motivation for reinforcement learning agents. In *Affective Computing and Intelligent Interaction: 4th International Conference, ACII 2011, Memphis, TN, USA, October 9–12, 2011, Proceedings, Part I 4*, 326–336. Springer.

Stein, M. B.; and Sareen, J. 2015. Generalized anxiety disorder. *New England Journal of Medicine*, 373(21): 2059–2068.

Sutton, R. S.; Barto, A. G.; et al. 1998. *Introduction to reinforcement learning*, volume 135. MIT press Cambridge.

Vaswani, A.; Shazeer, N.; Parmar, N.; Uszkoreit, J.; Jones, L.; Gomez, A. N.; Kaiser, Ł.; and Polosukhin, I. 2017. Attention is all you need. *Advances in neural information processing systems*, 30.

Wang, A.; Singh, A.; Michael, J.; Hill, F.; Levy, O.; and Bowman, S. R. 2018. GLUE: A multi-task benchmark and analysis platform for natural language understanding. *arXiv preprint arXiv:1804.07461*.

Zhao, J.; Wu, M.; Zhou, L.; Wang, X.; and Jia, J. 2022. Cognitive psychology-based artificial intelligence review. *Frontiers in Neuroscience*, 16: 1024316.